\pdfoutput=1
\documentclass{article}

\PassOptionsToPackage{numbers, compress}{natbib}

\usepackage[dblblindworkshop, final]{neurips_2026}
\workshoptitle{8th Robot Learning Workshop: Is Physical AI Going Zero-Shot?}

\usepackage[utf8]{inputenc}
\usepackage[T1]{fontenc}
\usepackage{hyperref}
\usepackage{url}
\usepackage{booktabs}
\usepackage{amsmath,amssymb,amsfonts}
\usepackage{graphicx}
\usepackage{microtype}
\usepackage{xcolor}
\usepackage{enumitem}

\newif\ifvideo\videotrue

\title{ValueFormer: Stage-Aware Value Labels for Online\\Failure Detection and VLA Post-Training}

\author{%
  Inkyu Sa \quad Konstantin Stulov \quad Rajat Bhageria \\
  Chef Robotics, San Francisco, CA, USA \\
  \texttt{\{inkyu, konstantin, rajat\}@chefrobotics.ai} \\
  \normalfont Video: \url{https://bit.ly/4i4kWTL}
}

\begin{document}

\maketitle

\begin{abstract}
Generalist vision-language-action (VLA) policies now transfer to new scenes with little task-specific training, but they still fail silently: from the action stream alone, a collapsing rollout looks much like one making clean progress, because imitation supplies no notion of progress. The cheap alternative, a terminal success / failure bit, is learnable in principle yet too sparse to say \emph{when} a rollout went wrong. We argue that the per-frame label is the hard part, and that it works better when continuous and non-monotonic: a score rather than a single success or failure, and one free to fall and then rise again as a rollout goes wrong and recovers. Such a label is cheap to obtain: a person marks only the stage at which an episode went wrong, and the per-frame values are computed from that. We present ValueFormer, a policy-agnostic causal transformer ($3.5$\,M parameters) over a frozen DINOv3 backbone, emitting two per-frame signals in one forward pass: a smooth Monte Carlo value for advantage estimation and a sharp binary value for online mistake detection. Failed episodes carry a stage-aware, success-then-decay return. A teleoperation station already records an intervention flag; reusing it lifts held-out detection from $0.38$ to $0.82$ average precision and flags $95\%$ of mistakes before the operator reacts. On a real bimanual sandwich station, feeding the critic back as a per-frame post-training weight raises task completion from $70\%$ to $85\%$ and removes a repeat-pick failure mode in our trials, although at $n{=}20$ that gain sits within noise.
\end{abstract}

\section{Introduction}
\label{sec:intro}

Bimanual manipulation in unstructured human environments, such as a restaurant sandwich station, is now within reach of modern VLA policies. Models from the $\pi_0$ family~\cite{black2024pi0,black2025pi05,amin2025pi06}, together with OpenVLA~\cite{kim2024openvla}, RDT-1B~\cite{liu2024rdt1b}, and GigaBrain~\cite{gigabrain2025}, have reduced months of engineering to a few hours of demonstration.

\begin{figure}[t]
\centering
\includegraphics[width=\linewidth]{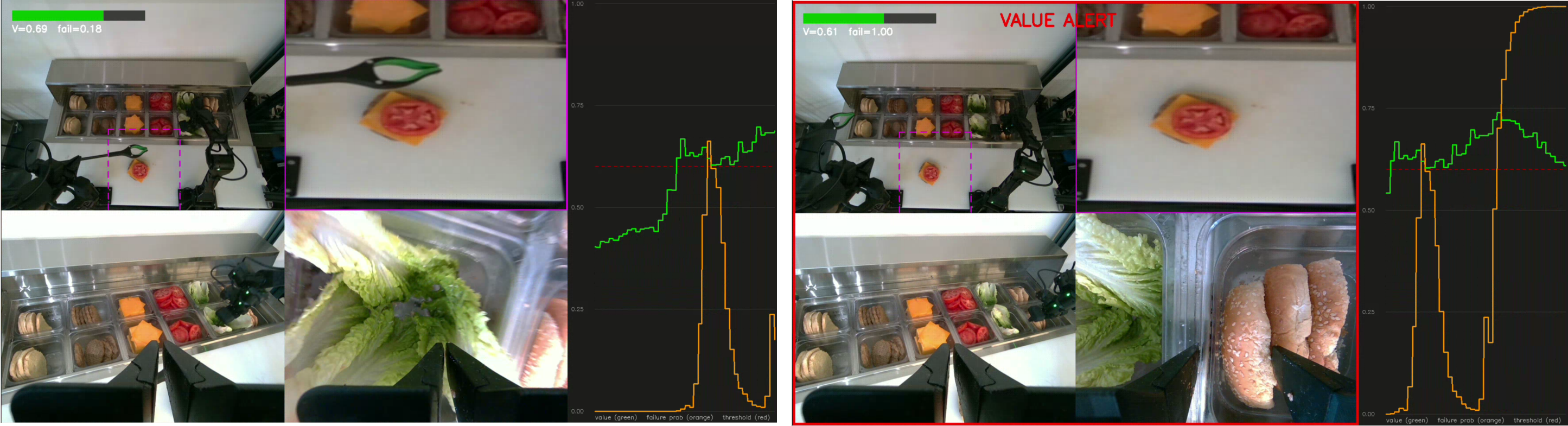}
\caption{ValueFormer live alongside a VLA policy on a held-out rollout: one forward pass emits a smooth progress value (green) and a sharp failure probability (orange). \emph{Left:} a healthy frame, no alert. \emph{Right:} after an external disturbance the lettuce is skipped, the failure probability saturates to $1.00$ and raises an operator-facing \textsc{value alert}, while the smooth value has barely moved\ifvideo\ (video: \url{https://bit.ly/4i4kWTL})\fi.}
\label{fig:teaser}
\end{figure}

However, a policy that transfers zero-shot to a new station has not been evaluated there, and if it was trained by imitation on successful demonstrations, it cannot report how well it is doing. In our sandwich station, the policy places six ingredients in order, one per stage. Whether a rollout is stuck on lettuce-leaf picking, about to place a second patty by mistake, or running correctly, it emits the same kinds of output: an action chunk, a video stream, and joint commands, none of which says whether it is on track. For an operator overseeing a fleet, that missing notion of \emph{progress} often decides whether one person can oversee many robots or each robot needs someone watching its video stream. Figure~\ref{fig:teaser} shows what we want instead: an alert at the frame a rollout goes wrong.

A value function would supply that signal: a scalar, read from the same observations, that tracks how close the episode is to completion. Actor-critic reinforcement learning gives this for free, but a flow-matching VLA trained by behavior cloning has no value baseline and no reward, and switching it over is impractical here: on-policy exploration on a live station is slow, wasteful of food, and unsafe unattended, and simulation is blocked by deformable food. We therefore learn the critic separately from the policy.

The question is then what supervision to ask for. Every rollout already carries the cheapest label imaginable: one bit at the end, success or failure. That bit is learnable in principle, but it cannot say \emph{when} things started to go wrong: one bit over a one-minute, multi-stage episode says nothing about which frames were responsible, the same sparsity that makes reinforcement learning expensive on real robots. It suits neither consumer: the critic needs an advantage, the \emph{difference} in value between nearby frames, which a single terminal label leaves undefined; the online mistake detector, a safety filter, needs the frame at which a mistake began. Label density alone is not enough either. A dense but \emph{wrongly shaped} label can be worse than no label, because the network fits it faithfully: the textbook Monte Carlo (MC) target sets the value to $0$ over every frame of a failed episode, which teaches the model that clean pre-failure observations predict failure. Two properties of the label shape follow. The target should be \emph{continuous}, a graded per-frame scalar rather than a terminal success or failure, and \emph{non-monotonic}, free to fall and rise again, because a rollout that drifts and then recovers runs good\,$\rightarrow$\,bad\,$\rightarrow$\,good whereas a progress clock only counts up. Neither property asks more of the annotator, which is the point of \emph{macro-labeling}: one stage index per failed episode and one interval per mistake expand into the per-frame continuum.

Getting that shape right is the central contribution here. Specifically, the contributions of this paper are:
\begin{itemize}[leftmargin=1.25em,itemsep=1pt,topsep=2pt,parsep=0pt]
  \item \textbf{Stage-aware, success-then-decay labeling.} A continuous, non-monotonic per-frame return expanded from one macro stage index: pre-failure frames of a failed episode follow the success curve and decay only afterwards. Of the five shapes trained under one fixed architecture it is the only one that neither reveals the outcome at $t{=}0$ nor jumps at the failure frame, and the label shape alone moves validation loss by $4\times$.
  \item \textbf{A causal dual-head architecture aligned with deployment.} One causal transformer over a frozen DINOv3 backbone emits a smooth progress value and a sharp per-frame failure signal in one forward pass. One frame cannot separate a mistake from ordinary periodic motion, so it reads an $8$\,s history of images and joint states; the triangular mask keeps training identical to inference.
  \item \textbf{Detection supervision that scales without new annotation.} The sharp head is supervised by mistake \emph{intervals}, so recovered mistakes also carry signal, and a teleoperation station's intervention flag already supplies those intervals for free: held-out detection rises from $0.38$ to $0.82$ average precision.
  \item \textbf{An on-robot A/B of the closed loop.} Two different critic-derived training weights each raise completion from $70\%$ to $85\%$ and remove a repeat-pick failure mode against a same-data flag-only baseline, within noise at $n{=}20$.
\end{itemize}
The remainder of this paper is structured as follows. Section~\ref{sec:method} describes the two per-frame targets and the model trained on them. We present our experimental results in Section~\ref{sec:exp}, and discuss what we think generalizes in Sections~\ref{sec:discussion} and~\ref{sec:conclusion}.

\paragraph{Related work.} Value and reward models learn progress from video (VIP~\cite{ma2022vip}, LIV~\cite{ma2023liv}, R3M~\cite{nair2022r3m}) or from reward-labeled experience (Q-Transformer~\cite{chebotar2023qtransformer}, IQL / CQL~\cite{kostrikov2021iql,kumar2020cql}), and VLAC~\cite{zhai2025vlac} folds critic and actor into one model. The closest work is $\pi^*_{0.6}$ / RECAP~\cite{amin2025pi06}: a critic trains on real-robot data, including on-policy rollouts and teleoperated interventions, then post-trains the policy that produced the data. Two things differ. Firstly, RECAP thresholds the advantage into a binary indicator that conditions the policy, whereas we keep its magnitude, so one continuous critic can supply both a soft advantage weight and a hard mask (Section~\ref{sec:exp:ab}), and it can be read per frame at deployment to raise the alert of Figure~\ref{fig:teaser}; RECAP's indicator is not read by an operator at run time. Secondly, RECAP's value counts steps to completion, so its targets are indexed by time; we index by task stage instead (Section~\ref{sec:method:labels}). Failure detectors for behavior-cloning policies (SAFE~\cite{gu2025safe}, AHA~\cite{duan2024aha}, I-FailSense~\cite{ifailsense2025}, StepEval~\cite{stepeval2025}) mostly emit an episode-level or per-subtask decision, which arrives only after the failure; hierarchical systems (Hi Robot~\cite{shi2025hirobot}, SayCan~\cite{ahn2022saycan}, HG-DAgger~\cite{kelly2019hgdagger}) route everything through a language supervisor with no calibrated progress signal.

\section{ValueFormer}
\label{sec:method}

\subsection{Stage-Aware, Success-Then-Decay Labels}
\label{sec:method:labels}

ValueFormer's two heads predict $V_\text{mc}(t)$ and $V_\text{bin}(t)$, both in $[0,1]$; this section defines their targets. Per-frame labels $v_k\in[0,1]$ for the smooth head come from an episode's outcome and, for failures, a stage-aware failure timing. Let $\gamma=0.99$ be the discount, $k\in\{0,\ldots,N{-}1\}$ the sample index at $2$\,Hz, and $N$ the number of sampled frames. Successful episodes follow a rising MC curve $v^{\text{succ}}_k=\gamma^{(N-1-k)}$, so $v_{N-1}=1$ and $v_0\approx0.23$ for a typical $73$\,s episode. For a failure, let $s_\text{fail}\in\{0,\ldots,N_s{-}1\}$ be the number of the $N_s$ stages completed before the mistake and $k_\text{fail}=\lfloor(s_\text{fail}/N_s)N\rfloor$ the frame where the failing stage begins, approximating every stage as an equal share of the episode. Then
\begin{equation}
v_k^{\text{fail}} =
\begin{cases}
\gamma^{(N-1-k)} & k \le k_\text{fail}~,\\[1pt]
v^{\text{succ}}_{k_\text{fail}}\,\gamma^{(k-k_\text{fail})} & k > k_\text{fail}~.
\end{cases}
\label{eq:label_fail}
\end{equation}

\begin{figure}[t]
\centering
\begin{minipage}[c]{0.73\linewidth}\centering
\includegraphics[width=\linewidth]{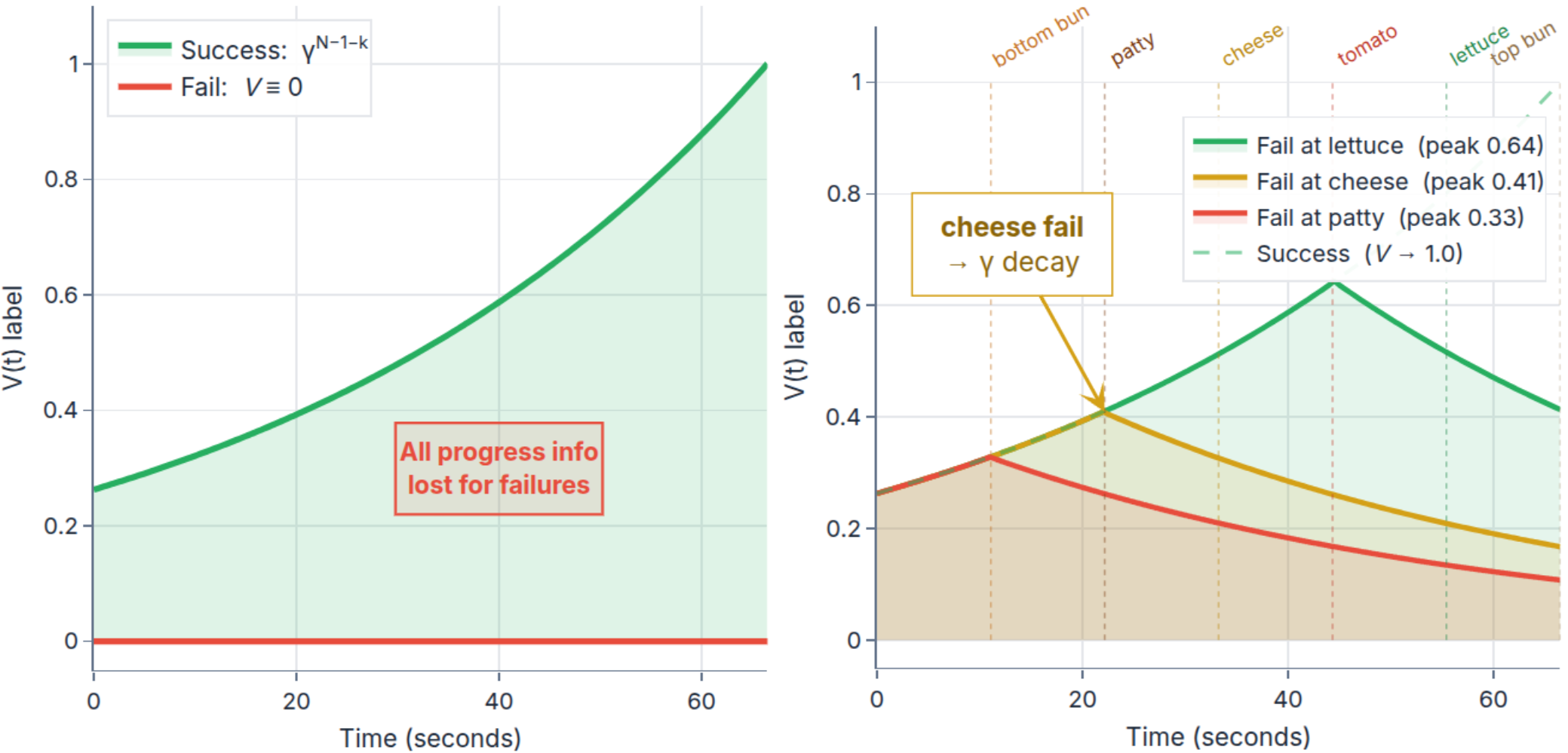}\\[1pt]
\makebox[0.5\linewidth]{\footnotesize (a)}\makebox[0.5\linewidth]{\footnotesize (b)}
\end{minipage}\hfill
\begin{minipage}[c]{0.23\linewidth}
\caption{Two targets for the frames of a failed episode. \textbf{(a)}~The naive Monte Carlo target flattens every frame to $0$. \textbf{(b)}~Eq.~(\ref{eq:label_fail}) follows the success curve to the failure stage and only then decays, so later failures peak higher (patty $0.33$, cheese $0.41$, lettuce $0.64$).}
\label{fig:stagelabels}
\end{minipage}
\end{figure}

Figure~\ref{fig:stagelabels} shows the effect of Eq.~(\ref{eq:label_fail}). The pre-failure prefix reuses the success curve because, up to $k_\text{fail}$, the observations are those of a working rollout with no visible cue of the outcome; a zero target over that prefix bleeds into clean-success inference at deployment. After $k_\text{fail}$ it decays at the same rate, a smooth target instead of a cliff. The exponential form matters downstream too: under a linear label the advantage $A_t=V_\text{mc}(t{+}H_A)-V_\text{mc}(t)$, for a lookahead horizon $H_A$, is constant over a success, so it cannot separate real progress from coasting.

\paragraph{A second, sharp target.} Eq.~(\ref{eq:label_fail}) changes by at most a factor of $\gamma$ between sampled frames, which is what a critic wants and the opposite of what a safety filter wants: a step at the moment a mistake begins, not a decaying tail. We therefore add a per-frame binary target $v^\text{bin}_k\in\{0,1\}$, built from hand-annotated mistake intervals $\{(t^{(j)}_\text{start},t^{(j)}_\text{end})\}$ on flagged rollouts: $v^\text{bin}_k=0$ inside any interval and $1$ elsewhere. An episode that succeeds overall then still contributes $0$-labeled frames over a transient mistake, such as a missed grasp the policy retries. Neither the naive MC target nor Eq.~(\ref{eq:label_fail}) expresses such a mistake: both are keyed to the episode outcome and assign $v^\text{succ}_k$ throughout a success.

\begin{figure}[t]
\centering
\includegraphics[width=0.95\linewidth]{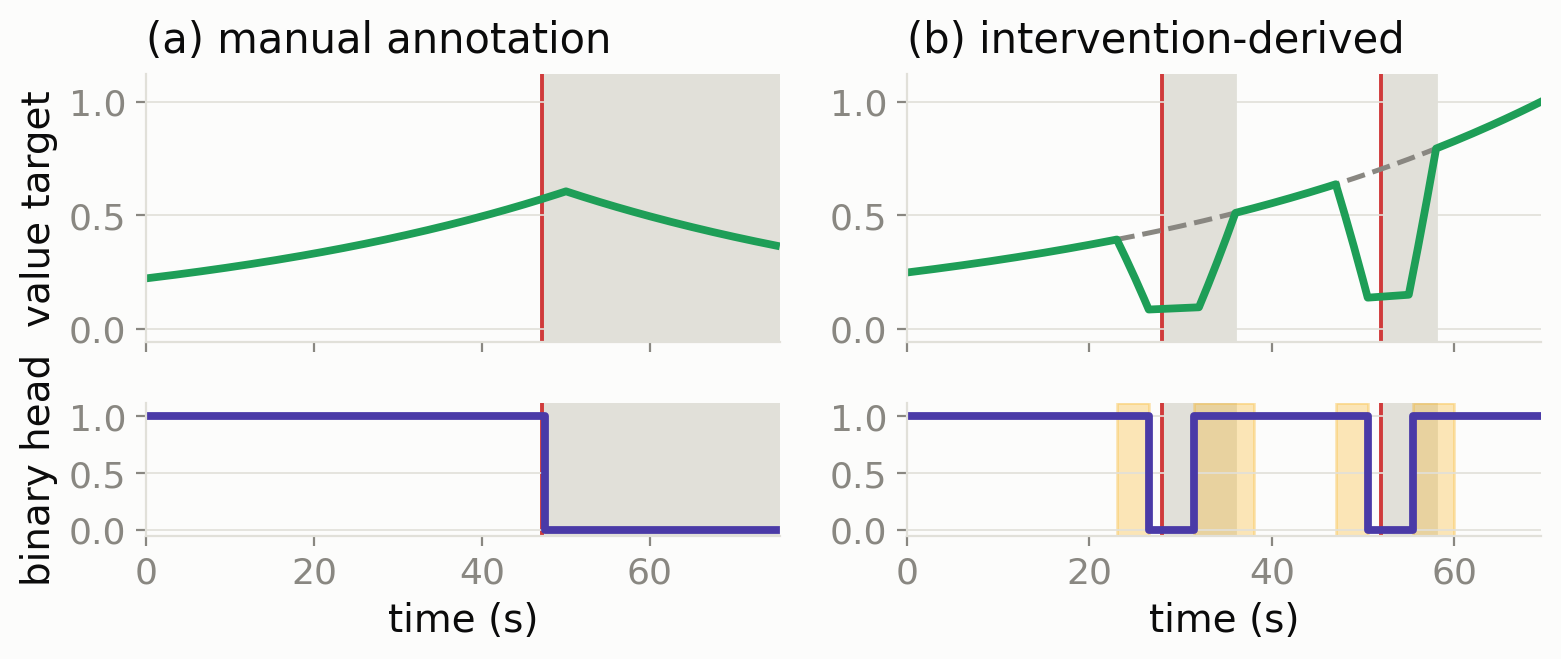}
\caption{Two routes to a per-frame label. Green is $V_\text{mc}$, purple $V_\text{bin}$, gray the mistake or takeover span with a red line at its onset, amber the frames excluded from the binary loss. \textbf{(a)}~A real annotated failure (``couldn't pick lettuce''): $V_\text{mc}$ is Eq.~(\ref{eq:label_fail}), $V_\text{bin}$ is $0$ over the marked interval. \textbf{(b)}~Two takeovers under the deployed labeler: $V_\text{mc}$ dips from the original success curve (dashed) and recovers by release, and $V_\text{bin}$ is $0$ only from $\lambda{=}1.5$\,s before the takeover onset to $3$\,s after it. Spans are illustrative.}
\label{fig:labels}
\end{figure}

\subsection{Architecture and Training}
\label{sec:method:arch}

Each frame is encoded from six views through one shared frozen DINOv3~\cite{oquab2024dinov3} ViT-L/16 at $518^2$: three camera images and three region-of-interest (ROI) crops. The encoder is never fine-tuned, and we did not test fine-tuning it: off-the-shelf DINOv3 features transfer to this station as they are. The six CLS tokens, the joint state and two time features are projected to $d_\text{model}{=}256$ and summed. A causal transformer ($L{=}2$, $H{=}4$) then runs over an $8$\,s window, and its last token feeds one MLP per head. Two properties drive that choice. Rollouts are periodic (arms leave and re-enter view, grippers open and close), so a single frame cannot separate a mistake from ordinary motion and the value needs a window of history. The triangular mask lets frame $t$ attend only to frames at or before $t$; a bidirectional encoder would not be admissible online. Both heads train jointly under binary cross-entropy (BCE). Appendix~\ref{app:arch} gives the parameter budget and the rest of the recipe.

\subsection{Intervention-Derived Segments: The Supervision That Scales}
\label{sec:method:hil}

Manual annotation does not scale: a fleet-scale loop produces more failures than a person can review. A human-in-the-loop (HIL) teleoperation station supplies the intervals for free. It logs a per-frame intervention flag, true while the operator is in control, and each unbroken run of that flag is one takeover: it starts when the operator judged the state bad ($t_\text{start}$) and ends when they handed a repaired state back ($t_\text{end}$). The flag cannot be used as a label directly, for two reasons. Firstly, it is a \emph{lagging} indicator: the operator reacts some time after the action that caused the problem, so we shift the $V_\text{bin}$ target a fixed lead $\lambda{=}1.5$\,s earlier than $t_\text{start}$ and \emph{exclude} the ambiguous run-up $[t_\text{start}-5\,\text{s},\,t_\text{start}-\lambda)$ from the binary loss. Secondly, the frames inside a takeover are the human's actions, not the policy's, so the value loss carries half weight from $t_\text{start}-\lambda$ to release, and the $V_\text{mc}$ target dips smoothly rather than stepping down (Appendix~\ref{app:arch}). Figure~\ref{fig:labels} shows both routes. We train on both sets, drawing only $20\%$ of frames from the noisier intervention-derived labels, and validate on the manual set.

\section{Experimental Results}
\label{sec:exp}

\paragraph{Setup.} All experiments run on a real bimanual sandwich station: two 7-DoF Trossen WidowX arms, six ingredient bins, and a $\pi_{0.5}$ policy~\cite{black2025pi05} prompted with ``make a sandwich''. The task has six ordered stages (bottom bun, patty, cheese, tomato, lettuce, top bun); a rollout succeeds if all six are placed correctly. The dataset is $1{,}427$ episodes in LeRobot v2 format~\cite{lerobot}, $213$k samples split by episode (Appendix~\ref{app:arch}). For the on-robot evaluation we grade each of a sandwich's six subtasks in $[0,1]$: $1.0$ is \emph{clean}, placed correctly with no retry, and near $0$ is \emph{catastrophic}, missed or damaged; values in between are partial credit (Appendix~\ref{app:traces}). A \emph{repeat-pick} is the policy picking an ingredient it has already placed.

\subsection{Ablation of the Fail-Episode Label Shape}
\label{sec:exp:ablation}

We trained one ValueFormer per label shape, varying only the value assigned to frames of failed episodes, under our earlier four-view single-head configuration so the dual head does not confound the label effect. Alongside Eq.~(\ref{eq:label_fail}) (MC-smooth) we consider four alternatives, all hard-zeroed after $k_\text{fail}$ and defined in Appendix~\ref{app:arch}: \textbf{A: outcome-scaled}, \textbf{B: cliff}, \textbf{C-linear} and \textbf{C-late}.

\begin{table}[t]
\centering
\caption{Held-out rollout metrics for the five fail-episode label shapes, with model and recipe held fixed. Validation BCE and mean absolute error (MAE) are each against the scheme's own target; $\bar V_\text{succ}$ and $\bar V_\text{fail}$ are mean predicted $V_\text{mc}$ over the success and failure subsets of the $21$ held-out rollouts.}
\label{tab:ablation}
\footnotesize
\begin{tabular}{lrrrrr}
\toprule
Scheme & Val loss & Rollout MAE & $\bar V_\text{succ}$ & $\bar V_\text{fail}$ & $\Delta V$ \\
\midrule
MC-smooth (ours)  & \textbf{0.00051} & \textbf{0.024} & 0.533 & 0.390 & 0.143 \\
A: outcome-scaled & 0.00192 & 0.058 & 0.504 & \textbf{0.195} & \textbf{0.309} \\
B: cliff          & 0.00202 & 0.047 & 0.529 & 0.275 & 0.254 \\
C-linear          & 0.00190 & 0.058 & 0.515 & 0.266 & 0.249 \\
C-late            & 0.00178 & 0.048 & 0.516 & 0.267 & 0.250 \\
\bottomrule
\end{tabular}
\end{table}

As a regression problem MC-smooth wins by $3.5\times$ on validation BCE and $2\times$ on rollout MAE against the nearest alternative (Table~\ref{tab:ablation}): its pre-failure prefix is identical to the success curve, so the network is not asked to map two visually similar prefixes to two different targets. Each scheme is scored against its own target, so these numbers compare learnability, not accuracy. As a binary detector MC-smooth fares worse in the $\Delta V$ column, and we argue that gap is a label artifact. Scheme A multiplies the whole pre-failure curve by $c$, which is fixed by how the episode ends, so $v^\text{fail}_0$ already differs from $v^\text{succ}_0$ before any visible cue: such a target is difficult to learn from an ongoing rollout. The hard-zeroed tails make a collapse after four correct ingredients look much like a failure at the bottom bun, whereas MC-smooth's tail carries how far the policy got. Scheme B has no leak at $t{=}0$ but asks the head to fit a step over an optically continuous transition, which surfaces as oscillation just before the cliff. Continuity is the main difference: MC-smooth alone stays graded across $k_\text{fail}$, peaking at the failure stage instead of collapsing to zero. The $4\times$ spread in validation BCE loss across these five shapes, with the model and recipe held fixed, is our main empirical claim.

\subsection{Detection from Intervention Flags}
\label{sec:exp:hil}

We test the intervention-derived labels on a separate HIL set: $134$ rollouts of the same task ($362$k frames, $333$ takeovers). We hold out $24$ episodes and compare two critics under an identical recipe, manual annotations alone versus manual plus intervention labels, at a matched $5\%$ per-frame false-positive rate. Detection \emph{lead} is the time from the first sustained alert to the takeover onset. Sustained means $1-V_\text{bin}$ holds above that rate's threshold over consecutive $0.5$\,s ticks rather than spiking for one; the onset comes from the raw flag, so it is label-independent.

Adding the intervention labels lifts average precision $0.38\to0.82$ and area under the ROC curve (AUROC) $0.69\to0.94$, takes the fraction of mistakes flagged before the operator reacts from $35\%$ to $95\%$, and doubles the median lead from $1.2$ to $2.4$\,s, at the same false-alarm budget (Table~\ref{tab:hil}). The gain is threshold-free, the curves dominating over their whole range, and robust: all $48$ manual$+$intervention runs beat all $6$ manual-only runs (Figure~\ref{fig:sweep}). Appendix~\ref{app:traces} traces the dual head through four held-out episodes. One caveat: this held-out ground truth is itself intervention-derived, so it measures agreement with \emph{where the operator intervened}, not an independent notion of failure. The lead time is the exception.

\subsection{On-Robot A/B: Critic-Derived Post-Training Weights}
\label{sec:exp:ab}

The critic's two outputs give several ways to weight the policy's own post-training data; we compare two against a flag-only control (\emph{intervention-weighted regression}, IWR: up-weight human-takeover frames, no critic). Every scheme floors takeover frames at $w{=}9$ and differs only on the \emph{autonomous} frames of takeover episodes, which the control leaves at $1$. Among those, \texttt{vf-mask} sets $w{=}0$ where $V_\text{bin}<0.5$ marks a failing frame, and \texttt{vf-awr} scales the weight by $\mathrm{clip}(e^{A_t/\beta},0.1,9)$ (Appendix~\ref{app:arch}). The last is one step of offline policy iteration rather than reweighted imitation: advantage-weighted regression with $V_\text{mc}$ as the critic.

\begin{table}[t]
\centering
\caption{On-robot A/B of two critic-derived weights against the flag-only IWR control ($20$ sandwiches each). $\uparrow$ higher is better, $\downarrow$ lower is better.}
\label{tab:ab}
\footnotesize
\begin{tabular}{lcccc}
\toprule
Metric & $\pi_{0.5}$ (no HIL) & IWR (control) & \texttt{vf-mask} & \texttt{vf-awr} \\
\midrule
Completion $\uparrow$              & $65\%$ ($13/20$) & $70\%$ ($14/20$)  & $\mathbf{85\%}$ ($17/20$) & $\mathbf{85\%}$ ($17/20$) \\
\quad Wilson $95\%$ CI             & $[43,82]$        & $[48,85]$         & $[64,95]$                 & $[64,95]$ \\
\quad Sandwich $2$ only $\uparrow$ & $50\%$           & $60\%$            & $\mathbf{90\%}$           & $80\%$ \\
Mean subtask $\uparrow$            & $0.88$           & $0.91$            & $0.94$                    & $\mathbf{0.96}$ \\
Clean subtasks $\uparrow$          & $82\%$           & $88\%$            & $88\%$                    & $\mathbf{92\%}$ \\
Catastrophic subtasks $\downarrow$ & $10$             & $7$               & $\mathbf{2}$              & $\mathbf{2}$ \\
Repeat-picks $\downarrow$          & $0$              & $4$               & $0$                       & $0$ \\
Median duration $\downarrow$       & $1{:}16$         & $\mathbf{1{:}15}$ & $1{:}21$                  & $1{:}22$ \\
\bottomrule
\end{tabular}
\end{table}

All three variants train on the same $985$k-frame set ($134$ HIL episodes of Section~\ref{sec:exp:hil} merged $2{:}1$ with replayed autonomous data), so each A/B isolates the critic-derived weight. ValueFormer supplies only the training weight and is not part of the policy at inference. Each variant is scored over $10$ trials of two consecutive sandwiches with no scene reset, so $20$ sandwiches and $120$ graded subtasks. One sandwich from a fresh station is already reliable, whereas drift makes the second a hard case.

Both critic weights reach the same headline (Table~\ref{tab:ab}), but they get there differently. \texttt{vf-mask} gives the flattest floor, lifting the worst subtasks rather than the best (tomato $0.82\to0.97$ and lettuce $0.82\to0.95$, where the control was soft), and is the only variant that improves on the harder second sandwich. \texttt{vf-awr} reaches the highest ceiling, with perfect tomato and lettuce subtasks and the best mean subtask score, at the cost of struggling more on cheese. The repeat-pick result is the clearest signal: that reflex survived every flag-only lever (more correction data, a doubled replay share, pause-frame removal) yet does not appear once under either critic weight, plausibly because the critic strips the low-value drift before a re-pick from the gradient while the takeover frames that teach the recovery \emph{motion} keep full weight.

We regard these results as suggestive rather than established. At $n{=}20$ the completion differences sit within noise (Fisher exact $p{=}0.45$), several headline rows track the same three extra completed sandwiches, the catastrophic-subtask count included, and the four columns come from separate sessions, so ingredient staging is uncontrolled across them. What supports the result is the isolation of the comparison and the agreement of two different critic-derived weights; a paired re-evaluation would supply the power it lacks.

\section{Discussion and Limitations}
\label{sec:discussion}

\paragraph{Two consumers, two targets.} One head failed to serve both consumers at every $\gamma$ we tried; separate targets over a shared backbone recover both regimes for one extra $33$k-parameter head (Table~\ref{tab:arch}).

\paragraph{Fitting the critic next to the policy.} Running it live pushes $\pi_{0.5}$ inference from $58.6$\,ms in isolation to a $p90$ of $100.6$\,ms, close to the $\approx\!100$\,ms budget left by the $30$\,Hz loop's three-step action-chunk prefetch, not the $33.3$\,ms of one control tick; a batched bf16 encode path cuts the critic's $2$\,Hz tick cost from $266$ to $\approx\!52$\,ms and restores the margin at no measurable cost in AUROC (Appendix~\ref{app:serving}).

\paragraph{Limitations.} Firstly, predictions lag the true failure by several seconds, roughly half the causal history window; an asymmetric attention kernel favoring recent frames may help. Secondly, failure-stage annotation for the smooth head is still manual: a Gemma-4-26B~\cite{gemma4} stage labeler reached only $0.21$ MAE on the failure stage as a fraction of the episode, above the $0.10$ we judged usable. Lastly, the closed-loop evidence covers one path at $n{=}20$ and within noise, leaving abort triggering and advantage \emph{conditioning} untested, and every result comes from one task on one station, with no external failure detector as a baseline, so the cross-task claims we would most like to make are unsupported.

\section{Conclusions}
\label{sec:conclusion}

ValueFormer overlays a calibrated per-frame progress signal on a $\pi_0$-family VLA; its technical core is the per-frame label, a stage-aware, success-then-decay return. What generalizes is the recipe rather than the curve: keep the annotation macro, and expand it into a per-frame target free to fall and recover. A second head, fed at fleet scale by the station's own intervention flag, lifts held-out detection from $0.38$ to $0.82$ average precision, and as a training weight the same critic raises completion from $70\%$ to $85\%$, within noise at $n{=}20$. Because it is a percent of the policy's size and trains in three minutes, one critic can follow a VLA across the generations that replace it; the stage decomposition and annotated failure set stay task-specific. As policies generalize to new stations, online failure detection should generalize with them.

\begin{ack}
The authors thank the Chef Robotics team and the open-source OpenPI, LeRobot, and DINOv3 communities, whose artifacts made this work possible. This work is supported by Chef Robotics' Sandi project.
\end{ack}

\bibliographystyle{plainnat}
\bibliography{references}

\begin{thebibliography}{23}
\providecommand{\natexlab}[1]{#1}
\providecommand{\url}[1]{\texttt{#1}}
\expandafter\ifx\csname urlstyle\endcsname\relax
  \providecommand{\doi}[1]{doi: #1}\else
  \providecommand{\doi}{doi: \begingroup \urlstyle{rm}\Url}\fi

\bibitem[Ahn et~al.(2022)Ahn, Brohan, Brown, Chebotar, Cortes, David, Finn,
  Gopalakrishnan, Hausman, Herzog, et~al.]{ahn2022saycan}
Michael Ahn, Anthony Brohan, Noah Brown, Yevgen Chebotar, Omar Cortes, Byron
  David, Chelsea Finn, Keerthana Gopalakrishnan, Karol Hausman, Alex Herzog,
  et~al.
\newblock Do as i can, not as i say: Grounding language in robotic affordances.
\newblock \emph{arXiv preprint arXiv:2204.01691}, 2022.

\bibitem[Amin et~al.(2025)Amin, Aniceto, Balakrishna, Black, Conley, Connors,
  Darpinian, Dhabalia, DiCarlo, Driess, Equi, Fang, Finn, Glossop, Godden,
  Goryachev, Groom, Hancock, Hausman, Hussein, Ichter, Jakubczak, Jen, Jones,
  Katz, Ke, Kuchi, Lamb, LeBlanc, Lu, Mano, Mothukuri, Pertsch, Ren, Rez, Shi,
  Smith, Springenberg, Stachowicz, Swerdlow, Tanner, Torne, Vuong, Walling,
  Wang, Williams, Yu, Zhilinsky, and Zhou]{amin2025pi06}
Ali Amin, Raichelle Aniceto, Ashwin Balakrishna, Kevin Black, Ken Conley, Grace
  Connors, James Darpinian, Karan Dhabalia, Jared DiCarlo, Danny Driess,
  Michael Equi, Yunhao Fang, Chelsea Finn, Catherine Glossop, Thomas Godden,
  Ivan Goryachev, Lachy Groom, Hunter Hancock, Karol Hausman, Gashon Hussein,
  Brian Ichter, Szymon Jakubczak, Rowan Jen, Tim Jones, Ben Katz, Liyiming Ke,
  Chandra Kuchi, Marinda Lamb, Devin LeBlanc, Yao Lu, Vishnu Mano, Mohith
  Mothukuri, Karl Pertsch, Allen~Z. Ren, Charvi Rez, Lucy~Xiaoyang Shi, Laura
  Smith, Jost~Tobias Springenberg, Kyle Stachowicz, Alex Swerdlow, James
  Tanner, Marcel Torne, Quan Vuong, Anna Walling, Haohuan Wang, Blake Williams,
  Lili Yu, Ury Zhilinsky, and Zhiyuan Zhou.
\newblock {$\pi^*_{0.6}$}: A {VLA} that learns from experience.
\newblock \emph{arXiv preprint arXiv:2511.14759}, 2025.

\bibitem[Black et~al.(2024)Black, Brown, Driess, Esmail, Equi, Finn, Fusai,
  Groom, Hausman, Ichter, Jakubczak, Jones, Ke, Levine, Li-Bell, Mothukuri,
  Nair, Pertsch, Shi, Tanner, Vuong, Walling, Wang, and
  Zhilinsky]{black2024pi0}
Kevin Black, Noah Brown, Danny Driess, Adnan Esmail, Michael Equi, Chelsea
  Finn, Niccolo Fusai, Lachy Groom, Karol Hausman, Brian Ichter, Szymon
  Jakubczak, Tim Jones, Liyiming Ke, Sergey Levine, Adrian Li-Bell, Mohith
  Mothukuri, Suraj Nair, Karl Pertsch, Lucy~Xiaoyang Shi, James Tanner, Quan
  Vuong, Anna Walling, Haohuan Wang, and Ury Zhilinsky.
\newblock {$\pi_0$}: A vision-language-action flow model for general robot
  control.
\newblock \emph{arXiv preprint arXiv:2410.24164}, 2024.

\bibitem[Black et~al.(2025)Black, Brown, Darpinian, Dhabalia, Driess, Esmail,
  Equi, Finn, Fusai, Galliker, Ghosh, Groom, Hausman, Ichter, Jakubczak,
  LeBlanc, Levine, Li-Bell, Mothukuri, Nair, Pertsch, Ren, Shi, Smith,
  Springenberg, Stachowicz, Tanner, Vuong, Walke, Walling, Wang, Yu, and
  Zhilinsky]{black2025pi05}
Kevin Black, Noah Brown, James Darpinian, Karan Dhabalia, Danny Driess, Adnan
  Esmail, Michael Equi, Chelsea Finn, Niccolo Fusai, Manuel~Y. Galliker, Dibya
  Ghosh, Lachy Groom, Karol Hausman, Brian Ichter, Szymon Jakubczak, Devin
  LeBlanc, Sergey Levine, Adrian Li-Bell, Mohith Mothukuri, Suraj Nair, Karl
  Pertsch, Allen~Z. Ren, Lucy~Xiaoyang Shi, Laura Smith, Jost~Tobias
  Springenberg, Kyle Stachowicz, James Tanner, Quan Vuong, Homer Walke, Anna
  Walling, Haohuan Wang, Lili Yu, and Ury Zhilinsky.
\newblock {$\pi_{0.5}$}: A vision-language-action model with open-world
  generalization.
\newblock \emph{arXiv preprint arXiv:2504.16054}, 2025.

\bibitem[Cadene et~al.(2024)Cadene, Alibert, Soare, Gallouedec, Zouitine, and
  Wolf]{lerobot}
Remi Cadene, Simon Alibert, Alexander Soare, Quentin Gallouedec, Adil Zouitine,
  and Thomas Wolf.
\newblock {LeRobot}: State-of-the-art machine learning for real-world robotics
  in {PyTorch}.
\newblock \url{https://github.com/huggingface/lerobot}, 2024.

\bibitem[Chebotar et~al.(2023)Chebotar, Vuong, Irpan, Hausman, Xia, Lu, Kumar,
  Yu, Herzog, Pertsch, et~al.]{chebotar2023qtransformer}
Yevgen Chebotar, Quan Vuong, Alex Irpan, Karol Hausman, Fei Xia, Yao Lu, Aviral
  Kumar, Tianhe Yu, Alexander Herzog, Karl Pertsch, et~al.
\newblock {Q-Transformer}: Scalable offline reinforcement learning via
  autoregressive {Q-Functions}.
\newblock In \emph{Conference on Robot Learning ({CoRL})}, 2023.

\bibitem[Duan et~al.(2025)Duan, Pumacay, Kumar, Wang, Tian, Yuan, Krishna, Fox,
  Mandlekar, and Guo]{duan2024aha}
Jiafei Duan, Wilbert Pumacay, Nishanth Kumar, Yi~Ru Wang, Shulin Tian, Wentao
  Yuan, Ranjay Krishna, Dieter Fox, Ajay Mandlekar, and Yijie Guo.
\newblock {AHA}: A vision-language-model for detecting and reasoning over
  failures in robotic manipulation.
\newblock In \emph{International Conference on Learning Representations
  ({ICLR})}, 2025.
\newblock arXiv:2410.00371.

\bibitem[ElMallah et~al.(2025)ElMallah, Chhajer, and Lee]{stepeval2025}
Ramy ElMallah, Krish Chhajer, and Chi-Guhn Lee.
\newblock Score the steps, not just the goal: {VLM}-based subgoal evaluation
  for robotic manipulation.
\newblock In \emph{{CoRL} Workshop on Evaluation and Deployment of Generalist
  Robot Policies}, 2025.
\newblock arXiv:2509.19524.

\bibitem[{Gemma Team}(2026)]{gemma4}
{Gemma Team}.
\newblock Gemma 4 technical report.
\newblock \emph{Google DeepMind technical report}, 2026.
\newblock Checkpoint: \texttt{google/gemma-4-26b-a4b}.

\bibitem[{GigaBrain Team}(2026)]{gigabrain2025}
{GigaBrain Team}.
\newblock {GigaBrain-0.5M$^*$}: A {VLA} that learns from world model-based
  reinforcement learning.
\newblock \emph{arXiv preprint arXiv:2602.12099}, 2026.

\bibitem[Grislain et~al.(2026)Grislain, Rahimi, Sigaud, and
  Chetouani]{ifailsense2025}
Cl{\'e}mence Grislain, Hamed Rahimi, Olivier Sigaud, and Mohamed Chetouani.
\newblock {I-FailSense}: Towards general robotic failure detection with
  vision-language models.
\newblock In \emph{{IEEE} International Conference on Robotics and Automation
  ({ICRA})}, 2026.
\newblock arXiv:2509.16072.

\bibitem[Gu et~al.(2025)Gu, Ju, Sun, Gilitschenski, Nishimura, Itkina, and
  Shkurti]{gu2025safe}
Qiao Gu, Yuanliang Ju, Shengxiang Sun, Igor Gilitschenski, Haruki Nishimura,
  Masha Itkina, and Florian Shkurti.
\newblock {SAFE}: Multitask failure detection for vision-language-action
  models.
\newblock In \emph{Advances in Neural Information Processing Systems
  ({NeurIPS})}, 2025.
\newblock arXiv:2506.09937.

\bibitem[Kelly et~al.(2019)Kelly, Sidrane, Driggs-Campbell, and
  Kochenderfer]{kelly2019hgdagger}
Michael Kelly, Chelsea Sidrane, Katherine Driggs-Campbell, and Mykel~J.
  Kochenderfer.
\newblock {HG-DAgger}: Interactive imitation learning with human experts.
\newblock In \emph{{IEEE} International Conference on Robotics and Automation
  ({ICRA})}, 2019.

\bibitem[Kim et~al.(2024)Kim, Pertsch, Karamcheti, Xiao, Balakrishna, Nair,
  Rafailov, Foster, Lam, Sanketi, et~al.]{kim2024openvla}
Moo~Jin Kim, Karl Pertsch, Siddharth Karamcheti, Ted Xiao, Ashwin Balakrishna,
  Suraj Nair, Rafael Rafailov, Ethan Foster, Grace Lam, Pannag Sanketi, et~al.
\newblock {OpenVLA}: An open-source vision-language-action model.
\newblock \emph{arXiv preprint arXiv:2406.09246}, 2024.

\bibitem[Kostrikov et~al.(2022)Kostrikov, Nair, and Levine]{kostrikov2021iql}
Ilya Kostrikov, Ashvin Nair, and Sergey Levine.
\newblock Offline reinforcement learning with implicit {Q-Learning}.
\newblock In \emph{International Conference on Learning Representations
  ({ICLR})}, 2022.

\bibitem[Kumar et~al.(2020)Kumar, Zhou, Tucker, and Levine]{kumar2020cql}
Aviral Kumar, Aurick Zhou, George Tucker, and Sergey Levine.
\newblock Conservative {Q-Learning} for offline reinforcement learning.
\newblock In \emph{Advances in Neural Information Processing Systems
  ({NeurIPS})}, 2020.

\bibitem[Liu et~al.(2024)Liu, Wu, Li, Tan, Chen, Wang, Xu, Su, and
  Zhu]{liu2024rdt1b}
Songming Liu, Lingxuan Wu, Bangguo Li, Hengkai Tan, Huayu Chen, Zhengyi Wang,
  Ke~Xu, Hang Su, and Jun Zhu.
\newblock {RDT-1B}: A diffusion foundation model for bimanual manipulation.
\newblock \emph{arXiv preprint arXiv:2410.07864}, 2024.

\bibitem[Ma et~al.(2023{\natexlab{a}})Ma, Kumar, Zhang, Bastani, and
  Jayaraman]{ma2023liv}
Yecheng~Jason Ma, Vikash Kumar, Amy Zhang, Osbert Bastani, and Dinesh
  Jayaraman.
\newblock {LIV}: Language-image representations and rewards for robotic
  control.
\newblock In \emph{International Conference on Machine Learning ({ICML})},
  2023{\natexlab{a}}.

\bibitem[Ma et~al.(2023{\natexlab{b}})Ma, Sodhani, Jayaraman, Bastani, Kumar,
  and Zhang]{ma2022vip}
Yecheng~Jason Ma, Shagun Sodhani, Dinesh Jayaraman, Osbert Bastani, Vikash
  Kumar, and Amy Zhang.
\newblock {VIP}: Towards universal visual reward and representation via
  value-implicit pre-training.
\newblock In \emph{International Conference on Learning Representations
  ({ICLR})}, 2023{\natexlab{b}}.

\bibitem[Nair et~al.(2022)Nair, Rajeswaran, Kumar, Finn, and
  Gupta]{nair2022r3m}
Suraj Nair, Aravind Rajeswaran, Vikash Kumar, Chelsea Finn, and Abhinav Gupta.
\newblock {R3M}: A universal visual representation for robot manipulation.
\newblock In \emph{Conference on Robot Learning ({CoRL})}, 2022.

\bibitem[Shi et~al.(2025)Shi, Ichter, Equi, Ke, Pertsch, Vuong, Tanner,
  Walling, Wang, Fusai, Li-Bell, Driess, Groom, Levine, and
  Finn]{shi2025hirobot}
Lucy~Xiaoyang Shi, Brian Ichter, Michael Equi, Liyiming Ke, Karl Pertsch, Quan
  Vuong, James Tanner, Anna Walling, Haohuan Wang, Niccolo Fusai, Adrian
  Li-Bell, Danny Driess, Lachy Groom, Sergey Levine, and Chelsea Finn.
\newblock {Hi Robot}: Open-ended instruction following with hierarchical
  vision-language-action models.
\newblock In \emph{Proceedings of the 42nd International Conference on Machine
  Learning ({ICML})}, 2025.

\bibitem[Sim{\'e}oni et~al.(2025)Sim{\'e}oni, Vo, Seitzer, Baldassarre, Oquab,
  Jose, Khalidov, Szafraniec, Yi, Ramamonjisoa, et~al.]{oquab2024dinov3}
Oriane Sim{\'e}oni, Huy~V. Vo, Maximilian Seitzer, Federico Baldassarre, Maxime
  Oquab, Cijo Jose, Vasil Khalidov, Marc Szafraniec, Seungeun Yi, Micha{\"e}l
  Ramamonjisoa, et~al.
\newblock {DINOv3}.
\newblock \emph{arXiv preprint arXiv:2508.10104}, 2025.

\bibitem[Zhai et~al.(2025)Zhai, Zhang, Zhang, Huang, Zhang, Zhou, Zhang, Liu,
  Lin, and Pang]{zhai2025vlac}
Shaopeng Zhai, Qi~Zhang, Tianyi Zhang, Fuxian Huang, Haoran Zhang, Ming Zhou,
  Shengzhe Zhang, Litao Liu, Sixu Lin, and Jiangmiao Pang.
\newblock A vision-language-action-critic model for robotic real-world
  reinforcement learning.
\newblock \emph{arXiv preprint arXiv:2509.15937}, 2025.

\end{thebibliography}

\newpage
\appendix

\section{Architecture Detail}
\label{app:arch}

Table~\ref{tab:arch} lists the parameter budget of the deployed dual-head configuration. The frozen DINOv3 ViT-L/16 encoder is shared across all six views and contributes no trainable parameters. The $16$-frame window is left-padded with the first frame when fewer than $16$ samples are available. We shrank the context from the $32$ samples ($16$\,s) used in early single-head experiments because online detection benefits from sharper transitions: a longer window smooths the value over the half-window scale, which suits $V_\text{mc}$ but masks the moment a mistake begins for $V_\text{bin}$. Layer count, head count and window length were selected by sweeping against the held-out set; window length was the most impactful, swept over $\{8,16,32,64\}$. Optimization uses AdamW ($\text{lr}=10^{-4}$, weight decay $0.05$, five warm-up epochs, cosine annealing to zero, batch $256$, dropout $0.2$) with a sampler balanced over the $\{$expert-success, expert-fail, rollout-success, rollout-fail$\}$ groups, for at most $100$ epochs with patience $15$. The detection loss carries a batch-wise positive weight $\alpha_+=\mathrm{clip}(n_0/n_1,0.1,10)$ against the $95\%/5\%$ class imbalance, where $n_0$ and $n_1$ are the per-batch counts of mistake and good frames; BCE on the value beats squared error after the sigmoid, which saturates near the $\{0,1\}$ endpoints where success and failure diverge most. The six views are three full camera images (top, left wrist, right wrist), a fixed ROI crop of the top-view gripper region, and two wrist-gripper ROI crops; a missed grasp shows up in a wrist crop several seconds before the top-down view registers the consequence. The per-view CLS tokens give a $6\times1024$ frame descriptor, and the three projections are combined by element-wise \emph{sum} (in early experiments concatenation let vision dominate). The dataset is $1{,}249$ teleoperated expert successes plus $178$ autonomous rollouts ($90$ success, $88$ failure), subsampled to $213{,}102$ samples, with $21$ further rollouts ($12$ success, $9$ failure) held out. The advantage weight uses $H_A{=}50$ frames and $\beta{=}0.3$. Over an intervention segment the $V_\text{mc}$ target is multiplied by a trapezoid: a linear ramp from $1$ down to a floor of $0.2$ over $[t_\text{start}-5\,\text{s},\,t_\text{start}-\lambda]$, that floor until halfway through the takeover, then a linear recovery to $1$ at release, with overlapping dips combined by minimum. One branch of the sweep behind Figure~\ref{fig:sweep} is sharper: it drives this floor to $0$ and lengthens the lead to $2$\,s and the run-up to $6$\,s. The manual $+$ intervention cluster dominates the manual-only one under both branches, so the detection result does not turn on the dip shape. The $0.5$ value-loss weight from $t_\text{start}-\lambda$ to release was fixed, not swept. With $c=s_\text{fail}/N_s$, the four alternative fail-episode shapes of Section~\ref{sec:exp:ablation} are \textbf{A}, $c\cdot v^\text{succ}_k$ before $k_\text{fail}$; \textbf{B}, the success curve before it; \textbf{C-linear}, decaying linearly from $v^\text{succ}_0$ to $c$; and \textbf{C-late}, the success curve until the last $25\%$ of the pre-failure window, then ramping to $c$. Under this configuration the smooth head reaches validation mean squared error (MSE) $3.0\times10^{-4}$, with mean $V_\text{mc}$ of $0.537$ on successes against $0.000$ on failures.

\begin{table}[h]
\centering
\caption{ValueFormer architecture summary for the dual-head ViT-L/16 setup with $(d_\text{model},L,H,\text{seq\_len})=(256,2,4,16)$ and six-view encoding.}
\label{tab:arch}
\footnotesize
\begin{tabular}{llr}
\toprule
Block & Definition (shape) & \# params \\
\midrule
Frozen encoder & 6 views, DINOv3 ViT-L/16 CLS concat $(B,S,6144)$ & frozen \\
\midrule
Vision projection & Linear$(6144\rightarrow256)$ & 1{,}573{,}120 \\
State projection & Linear$(14\rightarrow256)$ & 3{,}840 \\
Time projection & Linear$(2\rightarrow256)$ & 768 \\
Positional embedding & Learnable, 16 positions, dim 256 & 4{,}096 \\
\midrule
Causal transformer & 2 layers, 4 heads, GELU pre-norm, FFN $4d_\text{model}$ & 1{,}579{,}520 \\
Final normalization & LayerNorm$(256)$ on the last-token pathway & 512 \\
\midrule
$V_\text{mc}$ head & MLP $256\rightarrow512\rightarrow256\rightarrow1$ & 264{,}705 \\
$V_\text{bin}$ head & MLP $256\rightarrow128\rightarrow1$ & 33{,}025 \\
\midrule
Trainable total & projections $+$ transformer $+$ dual heads & 3{,}459{,}586 \\
\bottomrule
\end{tabular}
\end{table}

\section{Detection Results and Dual-Head Traces}
\label{app:traces}

\begin{figure}[h]
\centering
\includegraphics[width=0.82\linewidth]{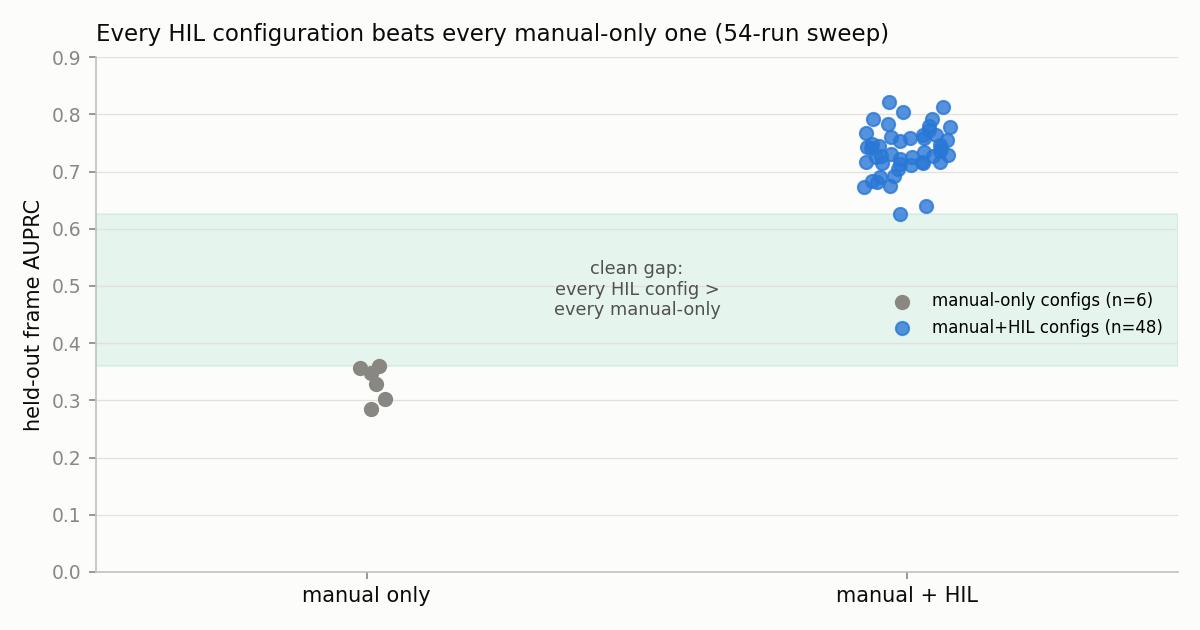}
\caption{The sweep behind the robustness claim of Section~\ref{sec:exp:hil}. Every manual $+$ intervention configuration ($48$ runs) beats every manual-only one ($6$ runs) on held-out average precision, with a clean gap between the two clusters. The effect survives variation in label shape, intervention share, context length, and detection-head weight.}
\label{fig:sweep}
\end{figure}

\begin{figure}[h]
\centering
\includegraphics[width=0.95\linewidth]{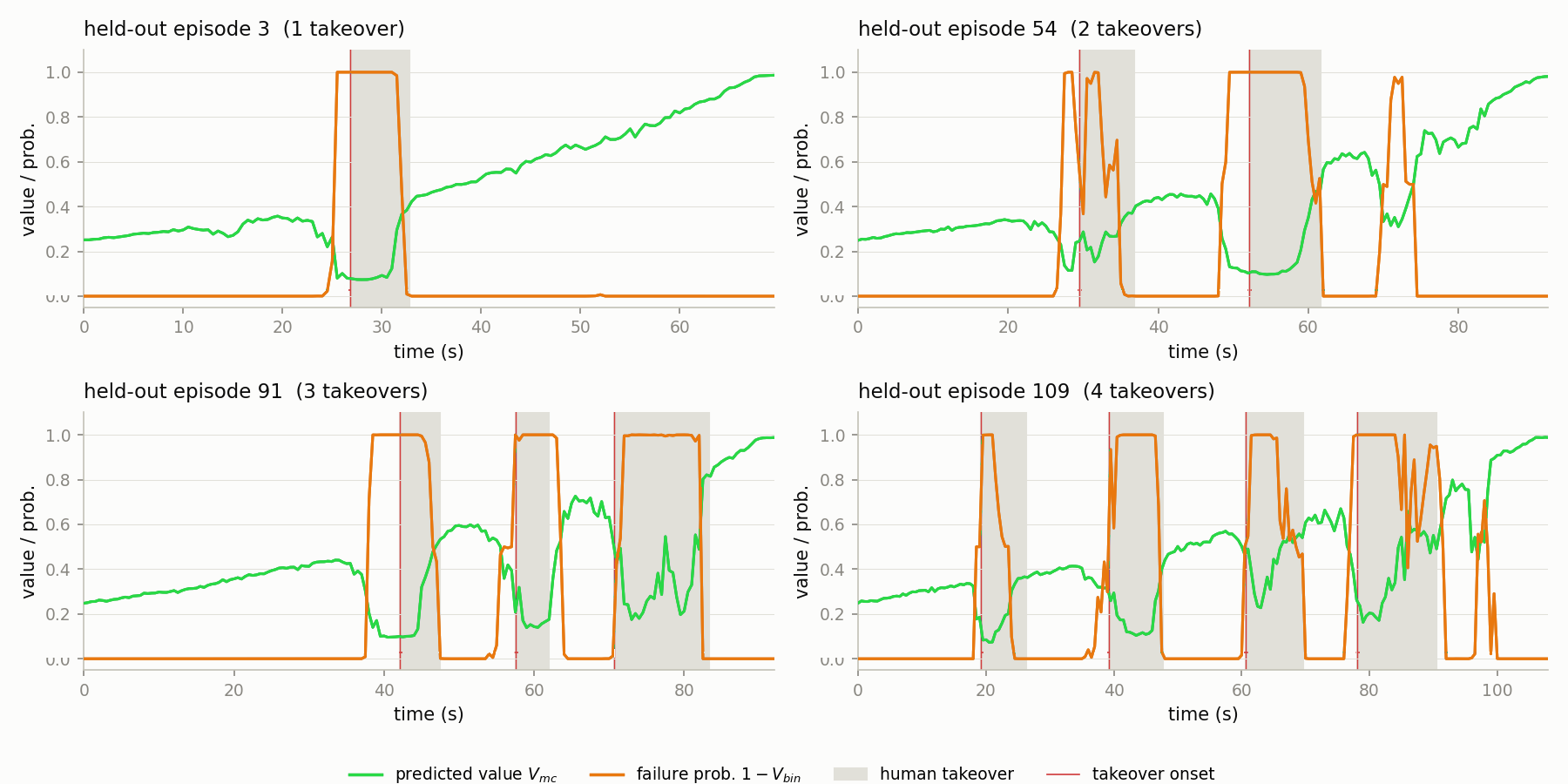}
\caption{The dual head on four held-out HIL episodes ($1$ to $4$ takeovers). Predicted value $V_\text{mc}$ (green) and failure probability $1-V_\text{bin}$ (orange); each shaded band is a human takeover and the red line marks its onset. The failure probability spikes into every takeover, while the value dips as the rollout drifts and recovers once the operator has repaired it.}
\label{fig:traces}
\end{figure}

\begin{figure}[h]
\centering
\includegraphics[width=\linewidth]{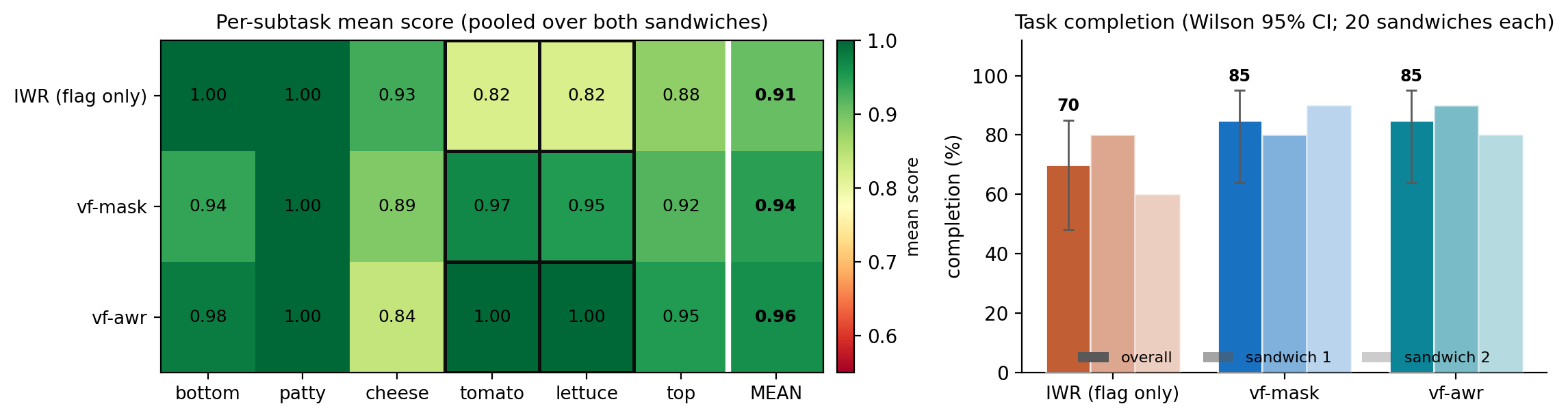}
\caption{The critic-derived weights against the flag-only IWR control. (Left)~per-subtask mean score pooled over both sandwiches, with a MEAN column; boxes mark the tomato and lettuce stages the control is soft on. (Right)~overall completion (Wilson $95\%$ CI) and the sandwich-1 / sandwich-2 split: both critic weights reach $85\%$, but \texttt{vf-mask} improves on the harder second sandwich ($60\%\to90\%$) while \texttt{vf-awr} dips ($90\%\to80\%$).}
\label{fig:ab}
\end{figure}

Table~\ref{tab:hil} gives the held-out detection numbers quoted in Section~\ref{sec:exp:hil}, Figure~\ref{fig:sweep} shows the sweep behind the robustness claim there, Figure~\ref{fig:traces} shows the dual head on four held-out HIL episodes, and Figure~\ref{fig:ab} breaks the on-robot A/B of Section~\ref{sec:exp:ab} down by subtask and by sandwich. Scores strictly between the \emph{clean} and \emph{catastrophic} endpoints of Section~\ref{sec:exp} do occur: a placement that needed a retry, for instance, takes partial credit. These intermediate grades follow no fixed ladder of allowed values; they are heuristic, read from the evaluation notes.

\begin{table}[h]
\centering
\caption{Held-out failure detection on the HIL set ($24$ episodes), manual annotations only versus manual $+$ intervention supervision, at a matched $5\%$ per-frame false-positive rate. The deployed model is a $2$-seed ensemble of the best configuration.}
\label{tab:hil}
\footnotesize
\begin{tabular}{lccc}
\toprule
       & Manual & \multicolumn{2}{c}{Manual $+$ intervention} \\
\cmidrule(lr){3-4}
Metric & only   & single & ensemble \\
\midrule
Average precision        & $0.38$   & $0.79$   & $\mathbf{0.82}$ \\
AUROC                    & $0.69$   & $0.93$   & $\mathbf{0.94}$ \\
Detected before takeover & $35\%$   & $95\%$   & $\mathbf{95\%}$ \\
Median lead              & $1.2$\,s & $2.4$\,s & $\mathbf{2.4}$\,s \\
Manual-val MSE           & $10^{-3}$ & $10^{-3}$ & $10^{-3}$ \\
\bottomrule
\end{tabular}
\end{table}

\section{Serving Profile}
\label{app:serving}

Table~\ref{tab:serving} gives the encode-path optimization headroom, measured back to back on the same model and inputs. Re-running the $2$-seed ensemble detector on bf16-encoded features of the $24$ held-out HIL episodes, at a matched $5\%$ false-positive rate, moves frame AUROC $0.943\to0.942$ and average precision $0.818\to0.816$, with the median detection lead unchanged at $2.4$\,s. Absolute times are inflated roughly $1.3\times$ because the policy server was resident during the run; the relative speedups are the takeaway. The batched FP32 output is numerically identical to the per-view path ($\max|\Delta|=3.5\times10^{-5}$), and bf16 embeddings stay cosine-aligned with FP32 (mean $0.99994$, min $0.99969$). Two orthogonal mitigations bound the residual interference without touching the model: pin the policy server to the memory fraction it actually uses (JAX preallocates $24$\,GB and uses under $7$), and cap the critic's compute share via CUDA MPS to bound the effect of a critic burst on the policy. The monitor is advisory and fail-safe, so a missed or late tick only delays an operator alert and does not enter the control path.

\begin{table}[h]
\centering
\caption{Value-encoder optimization headroom (encode path only, same model and inputs). Duty is the fraction of the $500$\,ms tick period the encode consumes.}
\label{tab:serving}
\footnotesize
\begin{tabular}{lrrr}
\toprule
Encode variant & Mean & Speedup & Duty @ $2$\,Hz \\
\midrule
Original: batch-1, FP32, host copies & $349.8$\,ms & $1.0\times$ & $70\%$ \\
Batched $\times6$, FP32, on-GPU      & $172.0$\,ms & $2.0\times$ & $34\%$ \\
Batched $\times6$, TF32              & $129.9$\,ms & $2.7\times$ & $26\%$ \\
Batched $\times6$, bf16 autocast     & $\mathbf{85.1}$\,ms & $\mathbf{4.1\times}$ & $\mathbf{17\%}$ \\
\bottomrule
\end{tabular}
\end{table}

\end{document}